\pdfoutput=1

\documentclass[11pt]{article}

\usepackage{emnlp2021}

\usepackage{times}
\usepackage{latexsym}

\usepackage[T1]{fontenc}

\usepackage[utf8]{inputenc}

\usepackage{microtype}
\usepackage{subfig}
\usepackage{graphicx}

\usepackage{todonotes}
\usepackage{multirow}
\graphicspath{ {./images/} }

\title{Residual Adapters for Parameter-Efficient ASR Adaptation to Atypical and Accented Speech}

\author{Katrin Tomanek\thanks{\quad Equal contribution}, Vicky Zayats\footnotemark[1], Dirk Padfield, Kara Vaillancourt, Fadi Biadsy \\
  Google, USA \\
  \texttt{\{katrintomanek,vzayats,padfield,kvailla,biadsy\}@google.com}
 }

\begin{document}
\maketitle
\begin{abstract}

Automatic Speech Recognition (ASR) systems are often optimized to work best for speakers with 
canonical speech patterns. Unfortunately, these systems perform poorly when tested on atypical speech
and heavily accented speech. It has previously been shown that personalization through model fine-tuning substantially improves performance. However, 
maintaining such large models per speaker is costly and difficult to scale.  We show that by adding a relatively small number of extra parameters to the encoder layers via so-called residual adapter, we can achieve similar adaptation gains compared to model fine-tuning, while only updating a tiny fraction (less than 0.5\%) of the model parameters.
We demonstrate this on two speech adaptation tasks (atypical and accented speech) and for two state-of-the-art ASR architectures.

\end{abstract}

\section{Introduction}
\label{sec:intro}

Automatic Speech Recognition (ASR) systems have achieved great success on a diverse set of acoustic and linguistic conditions, domains and speech patterns. 
State-of-the-art ASR systems are typically trained on tens of thousands of hours of speech data, and they perform well as long as these domains and conditions are well represented in the training data.  

Understandably, the distribution of such data typically focuses on the canonical {\em and} typical spoken language patterns of the target language, i.e., regional dialects, common accents and frequent non-native accents.
As a result, these systems may perform poorly on the tail of the distribution which may include ``heavily'' accented speech and/or speech with atypical speech patterns~\cite{darley1975motor}. Atypical speech includes dysarthric speech, speech impairments (due to, for example, ALS, stroke, traumatic brain injury, down syndrome, cerebral palsy, and MS), stuttering, deaf speech, or severe hyper-nasality due to cleft lip and palate. The lack of sufficient training data for these accents and atypical speech in the training distribution may result in a poor experience for a large segment of the population, leaving the less fortunate communities behind when it comes to speech-enabled technologies \cite{moore2018}.

Studies on accented speech showed word error rates (WER) twice or three times as high for accented speech compared to the more standard US accent \cite{sainath2020streaming, ghorbani2019leveraging}. 
Even worse performance is observed for speakers with speech impairments~\cite{moore2018}. Our goal, in this paper, is to efficiently build scalable models that can adapt to non-canonical or atypical speech.

It has been shown that speech models originally developed for typical speech can be successfully fine-tuned with limited amounts of data to accented or impaired speech~\cite{Zhu2019,Shor2019,Gale2019,mustafa2014,parrotron,extending_parrotron,green2021}. 
Nevertheless, one major challenge with adapting models to either individuals or small groups of speakers is that it is necessary to scale the number of models that need to be maintained and hosted. For example, for smart devices powerful enough to run ASR models on-device, having to deploy and store an additional (potentially large) model may take up valuable on-device resources. Similarly, providing personalized models for a large population of speakers in a centralized/server-based scenario is not feasible.

We propose to mitigate this issue by injecting residual adapter layers into the architecture. Particularly, we use a bottleneck architecture that requires a tiny number of parameters ($<0.5\%$ in our scenario) compared to the full model update via fine-tuning.
Then, while keeping the original pre-trained model parameters frozen, we update only the parameters of the adapter layers as we train on the custom data of interest.
This provides an easy way to deploy and store adapted models: a (generic) base model is deployed to all clients, and each individual or group can receive a personalized set of trained adapter layers that is small in size.

The main contributions of the paper are as follows. We show that residual adapters work extremely well for acoustic adaptation of different speech models. We present extensive experiments with adapter layers in two very different ASR use-cases: personalized models for atypical speech, and group models for accented speech. We also demonstrate that adapter layers work well in two different, state-of-the-art end-to-end ASR architectures, Neural Network Transducers (RNN-T), and Transformer Transducers (T-T). This emphasizes the flexibility of this approach and its suitability as a standard alternative to full fine-tuning for arbitrary models.
Our results clearly demonstrate how adaptation via adapter layers solves the issue of parameter inefficiency while largely retaining the significant adaptation gains achievable through model adaptation with in-domain data.
\section{Related Work}
\label{sec:related}

Model fine-tuning
has been successfully applied to domain adaptation for a variety of NLP tasks~\cite{devlin2018bert, sun2019fine}, Machine Translation~\cite{freitag2016}, and speech recognition and conversation systems, including accented and atypical speech~\cite{Zhu2019,Shor2019,Gale2019,parrotron,extending_parrotron,green2021}

A major disadvantage of model fine-tuning is its parameter inefficiency since it retrains all (or a large portion of) the model parameters on given task- or domain-specific data, resulting in a copy of the model for that task/domain. This is especially problematic for personalization of models due to the resulting high number of specialized models.

Concatenating input features and speaker-dependent vectors, such as i-vectors, is a parameter-efficient speaker adaptive approach that has been applied to both acoustic models as well as end-to-end ASR models \cite{saon2013, saon2021}. However, only moderate improvements have been achieved, even on typical speech. We speculate that such a {\em static}, low dimensional representation may not be sufficient to capture the complex acoustic-phonetic patterns (e.g., consonant dropping and vowel dropping, extreme vowel reduction or lengthening, missing phonemes and even syllables, very irregular speaking rate and rhythm) often found in impaired speech. 

Residual adapters were originally introduced by \citet{rebuffi_2017} for computer vision tasks as an alternative to fine-tuning. These first residual adapter modules consisted of  a single projection layer added between layers of a pre-trained network. \citet{houlsby2019parameter} proposed a variation  consisting of a bottleneck structure (down-projection through feed forward layer, RELU, up-projection) for task-specific adaptation of BERT models. Adapter modules were added after each sub-layer within a transformer layer, and the weights of the residual adapters as well as existing layer normalization parameters were updated during training. Finally, ~\citet{bapna2019} have formulated a simplification of residual adapters in the context of domain-adaptation for Machine Translation. Each residual adapter module has its own layer normalization block, followed by a down- and up-projection feed forward network. They argued that by including layer normalization in the residual adapter block, these modules are plug-able into arbitrary blocks of pre-trained modules because they learn the activation pattern of the layer into which they are injected. 

\citet{kannan2019large} proposed to use adapters on top of multi-lingual ASR models to further improve their performance (they report up to 9\% WER improvement for some of the 5 languages of the multi-lingual model). Our focus is different in that we consider residual adapters in speech personalization scenarios where the number of adapted models is several orders of magnitudes higher (e.g., tens of thousand of speakers with atypical speech and potentially hundreds of accents and dialects) and also not static (e.g. speech impairments often progress over time).

Learning Hidden Unit Contribution (LHCU) \cite{swietojanski_2016} is another approach to more parameter efficient speaker adaptation.
Instead of updating all weights of a model, LHUC adds learned factors to the output of each hidden unit modulating their amplitude. However, \citet{bapna2019} have shown that using residual adapters is much more effective.
\section{Methods}
\label{sec:methods}

\begin{figure*}[t]
    \centering
    \subfloat[RNN-T/T-T architecture.]
    {\includegraphics[width=2.9in]{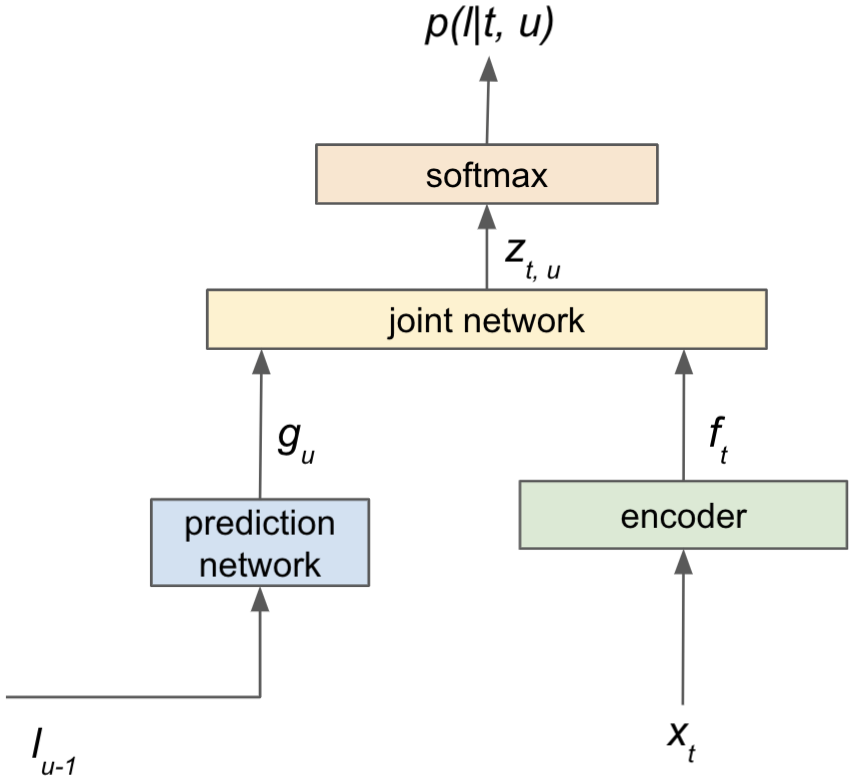}
    \label{fig:rnnt_arch}}\quad \quad
	\subfloat[Residual adapter and its integration into a Transformer encoder layer.]
	{\includegraphics[width=2.9in]{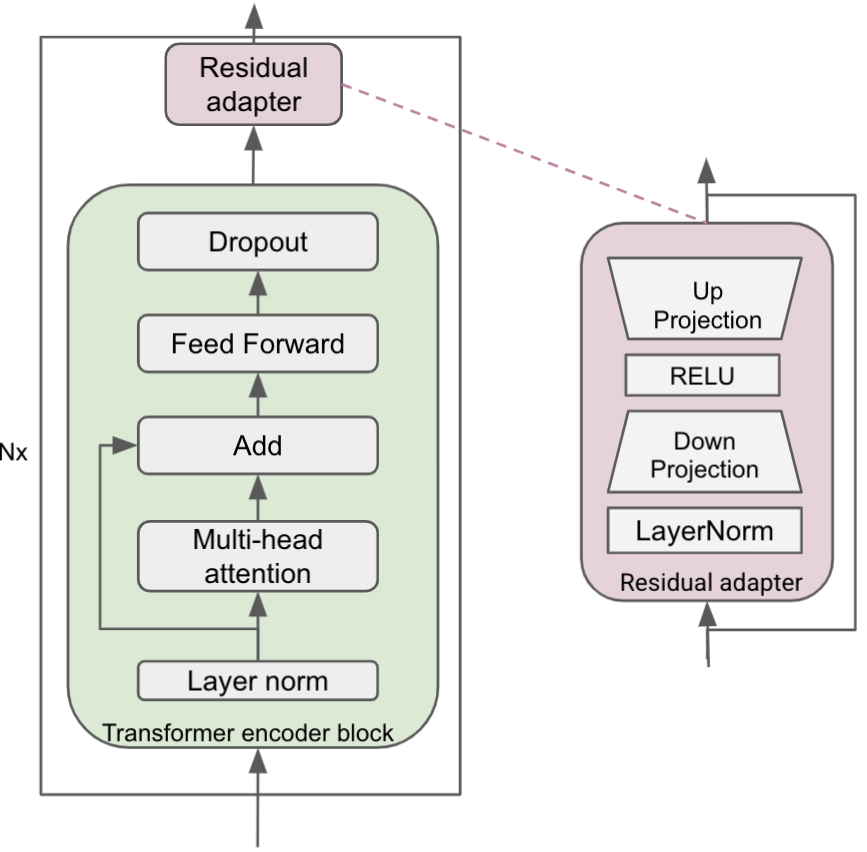}
	\label{fig:adapter}}\\
	\caption{Overview of RNN-T and T-T architectures and residual adapter module.}
\end{figure*}

For our experiments, we chose two state-of-the-art end-to-end ASR architectures: the Recurrent Neural Network Transducers (RNN-T) \cite{graves2013sequence,he2019streaming,sainath2020streaming} and the Transformer Transducer (T-T) \cite{zhang2020transformer}. Both architectures enable deployment on mobile devices, support streaming \cite{he2019streaming}, and have demonstrated high performance.

Both architectures consist of three main components: an encoder, a prediction network that incorporates label history and serves as a language model component (decoder), and a joint layer that combines predictions made by the encoder and the prediction network and feeds into a softmax.
All components of the two architectures are identical except the encoder stack. The prediction network consists of 2 uni-directional LSTM layers. Inputs are 128-dimensional log Mel features computed every 10 milliseconds. 4 consecutive features are stacked with a stride of 3 frames to yield a 512-dimensional input to the encoder every 30 milliseconds. 
Our output vocabulary consists of 4096 word piece tokens. Figure~\ref{fig:rnnt_arch} shows a high-level overview of both architectures.

For RNN-T, the encoder consists of 8 LSTM layers; for T-T, we use 15 Transformer layers in the encoder. Both architectures are trained with the RNN-T loss \cite{bagby2018efficient}.
To make T-T streamable, the attention calculation pays attention to past contexts only, which makes this architecture analogous to a uni-directional RNN. 

We propose to utilize residual adapter modules as outlined by \citet{bapna2019} for our adaptation approach. Each residual adapter block starts with layer normalization applied to the inputs, followed by a feed-forward layer with down-projection to dimension $d_b$, a non-linear activation (RELU), and another feed-forward layer with up-projection to the original input dimension $d_i$.
All weights of the residual adapter module are randomly initialized.

Figure~\ref{fig:adapter} shows such a residual adapter module and its integration within the Transformer encoder.
We add residual adapters to each encoder layer, resulting in 8 adapter layers for RNN-T and 15 adapter layers for T-T.
The bottleneck dimension $d_b$ enables control of the number of parameters of each residual adapter module and thus the capacity available during adaptation.
\section{Experiments}
\label{sec:experiments}

We analyze the performance of residual adapters as an alternative to model fine-tuning in two scenarios: adaptation to (a) atypical speech and (b) accented speech. 
For atypical speech, we build a personalized, speaker-dependent model for each speaker based on their data. For accented speech, we build per-accent models (i.e. speaker-independent models) and also experiment with a multi-accent adaptation scenario where one model is used for all covered accents.
We conduct experiments using both ASR transducer architectures (RNN-T and T-T).

\subsection{Accented Speech Dataset}
\label{sec:accent_data}
For the accented speech adaptation task, we use Mozilla's Common Voice corpus (v5.1) \cite{ardila2019common}. It contains spoken utterances of users reading sentences. Recordings were verified by other contributors using a simple voting system. While the full corpus contains 60 languages, for this work we use a subset containing only English recordings. 
We make use of Common Voice's metadata to extract accent information and use all 10 accents with more than 1k recordings, including (in order of decreasing number of recordings): England (\texttt{en}), India (\texttt{in}), Australia (\texttt{au}), Canada (\texttt{ca}), Scotland (\texttt{sc}), Ireland (\texttt{ir}), New Zealand (\texttt{nz}), Africa (\texttt{af}), Singapore (\texttt{si}), and Philippines (\texttt{ph}).

We randomly split all utterances from each accent into train/dev/test subsets. The resulting subset sizes per accent are shown in Table~\ref{tab:accents_both}. Table~\ref{tab:datasets} shows utterance counts and length (in words and seconds) aggregated across all accents.

\subsection{Atypical Speech Dataset}

We use the Euphonia corpus~\cite{macdonald2021} for the atypical speech personalization task. This corpus consists of over 1 million utterance recordings of over 1000 anonymized speakers with different types and severity levels of speech impairments. Similar to the Common Voice corpus, all recordings in the Euphonia corpus are prompted speech.
All our experiments are performed on a random subset of 100 speakers who have each recorded more than 1000 utterances. The resulting subset is very diverse, covering speakers with 15 different etiologies (31\% with amyotrophic lateral sclerosis (ALS), 20\% Down Syndrome, 14\% cerebral palsy, 6\% Parkinson's Disease, 5\% hearing impairment etc) and different speech impairment severity levels (47\% mild, 32\% moderate, 21\% severe).
We use the predefined per-speaker train, dev, and test splits (80\%/10\%/10\%).

Table~\ref{tab:datasets} shows utterance counts and length (in words and seconds) aggregated across all 100 speakers. Note that speakers with a speech impairment often have a lower speaking rate and frequently pause between individual words and before speaking. This is reflected in the relatively low ratio of words per second in the Euphonia corpus.

\begin{table*}[thb]
  \centering
  \begin{tabular}{|l|l|l|l|l|l|} 
  \hline
  Dataset & Subset & Hours & \# utts & Words per utt & Seconds per utt \\
  & & & & mean (std) & mean (std) \\
  \hline
  Euphonia & home automation & 73  & 69.3k  & 3.2 (1.4) & 3.8 (2.0) \\
           &  conversational & 67  & 37.1k  & 7.4 (4.0) & 6.5 (4.0) \\
  \hline
  Common Voice & 10 accents & 183 & 118.6k & 10.3 (2.8) & 5.6 (1.6) \\
  \hline
  \end{tabular}
  \caption{Utterance length statistics in number of words and seconds per dataset.}
  \label{tab:datasets}
\end{table*}

\subsection{Experimental Settings}

We follow a similar fine-tuning recipe as described in \citet{green2021}. We start from a speaker-independent base model pre-trained on 162k hours of typical (mostly American English) speech. This base model has been optimized to (a) be robust across various application domains and acoustic conditions, and (b) generalize well to unseen conditions \cite{narayanan2019}. The same base model is used across all of our experiments.

We use SpecAugment \cite{park2019} for data augmentation, limit training to a maximum of 50k steps (atypical speech) and 30k steps (accented speech) and employ small batch sizes (32 for atypical speech, 256 for accented speech with RNN-T, and 128 for accented speech with T-T). 

We only update the weights of the encoder layers, as our focus is on learning acoustic-phonetic variability as opposed to vocabulary and language variability. Accordingly, weights of the joint layer and the prediction network are always kept frozen.
When training with residual adapters, we freeze \emph{all} parameters of the base model and only update the residual adapter layers.
Table~\ref{tab:summary_results} shows the resulting number of parameters updated for the different adaptation strategies. For example, residual adapters with a bottleneck dimension of 16 yield more than $100\times$ parameter reduction compared to the encoder fine-tuning scenario.

Word error rate (WER) is measured on the respective test splits. 
The best checkpoints are chosen based on the WER on the dev split.

\section{Results}

\begin{table*}[htbp]
  \centering
  \begin{tabular}{|l|l|r|r|r|r|r|r|} \hline
  \multirow{2}{*}{Arch} & \multirow{2}{*}{Adaptation style} & \multicolumn{2}{|c|}{Updated params} & \multicolumn{2}{|c|}{Atypical Speech} & \multicolumn{2}{|c|}{Accented Speech} \\ \cline{3-8}
  & & Total & Relative & WER & Relative & WER & Relative \\
  \hline
  \hline  
  \multirow{5}{0cm}{\rotatebox{90}{RNN-T}} & Unadapted & -- & -- & 35.6 & -- & 19.9 & -- \\
  & Fine-tune, full enc & 98.7M & 81\% &
  6.0 & 80\% & 13.9 & 29\% \\
  & Fine-tune, enc layer 1  & 10.8M & 9\% & 10.9 & 61\% & 15.8 & 18\% \\
  & Fine-tune, enc layers 1-3 & 39.6M & 32\% & 6.9 & 75\% & 13.4 & 28\% \\
  & Residual Adapters, $b_d=16$ & 197K & <0.2\% & 6.8 & 77\% & 14.1 & 24\% \\
  \hline  
  \multirow{5}{0cm}{\rotatebox{90}{T-T}} & Unadapted & -- & -- & 38.4 & -- & 21.6 & -- \\
  & Fine-tune, full enc & 144.3M & 85\% & 6.1 & 78\% & 13.2 & 35\% \\
  & Fine-tune, enc layer 1 & 9.6M & 6\% & 10.8 & 60\% & 16.3 & 22\% \\
  & Fine-tune, enc layers 1-3 & 28.9M & 17\% & 8.4 & 72\% & 14.8 & 29\% \\
  & Residual Adapters, $b_d=16$  & 507K & <0.5\% & 7.1 & 75\% & 14.1 & 31\% \\
  \hline
  \end{tabular}
  \caption{
  Aggregated overview of adaptation results for both tasks. 
  The number of updated parameters is given as well as the percentage of the total; the total number of parameters is about 122M for RNN-T and about 168M for T-T.
  For atypical speech, we report median WER across all 100 speakers.  For accented speech, we report mean WER across 10 accents for the per-accent adaptation scenario. 
  The percentages in the WER columns are the relative WER improvement $\gamma$ (Eq. \ref{eq:gamma}) over the unadapted model. 
  }
  \label{tab:summary_results}
\end{table*}

\begin{figure}[t]
    \centering
    \includegraphics[width=0.47\textwidth]{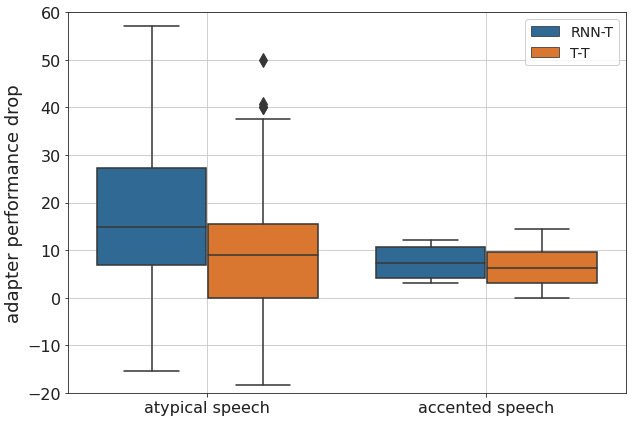}
    \caption{Distribution of adapter performance drop across all speakers/accents.}
    \label{fig:adapter_loss_combined}
\end{figure}

Table~\ref{tab:summary_results} compares fine-tuning versus residual adapters across both tasks and architectures.\footnote{The performance of RNN-T models is slightly better than T-T across all of our experiments despite the bigger encoder size of T-T. However, T-T's training time is much shorter.} Unless otherwise specified, we report per-accent (as opposed to multi-accent) adaptation results.
Adaptation performance is compared with performance on the \emph{unadapted} base model. In addition to WERs, we also report the \textit{relative WER improvement} over the unadapted model:
\begin{equation}
\label{eq:gamma}
    \gamma = \frac{\textrm{WER}(\textrm{unadapted}) - \textrm{WER}(\textrm{adapted})}{\textrm{WER}(\textrm{unadapted})}
\end{equation}

For residual adapters, we identified the best learning rate and bottleneck dimensions during hyper-parameter tuning on the dev set (see Section~\ref{ss:hparam_tuning}). In addition to comparing residual adapters to a scenario where we fine-tune the entire encoder, we also test the impact of fine-tuning only a few layers (1-3) of the encoder. However, this alternative for reducing the number of updated parameters is less efficient than residual adapters, which have a much lower parameter footprint due to their bottleneck architecture.

Table~\ref{tab:summary_results} shows that on the atypical speech personalization task
adapting the {\em full} encoder per speaker, we observe a relative reduction of 80\% in median WER across speakers for RNN-T.\footnote{\citet{green2021} report similar improvements over 500 speakers of the Euphonia corpus; their in-depth analysis shows that this holds across different severities and types of speech impairment.} However, this strategy requires 81\% of the model parameters to be updated and stored per speaker. Using residual adapters, on the other hand, we achieve relative WER reduction of 77\% across all speakers for RNN-T. Although fine-tuning is slightly better than adaptation with residual adapter layers, the latter only needs to update about $0.2\%$ of the parameters. We observe similar trends with T-T.

Comparing to a scenario where we update only a few bottom layers of the encoder, we observe a significant\footnote{Throughout this paper, we use paired t-tests to measure statistical significance (indicated as significant for p-values < 0.05)} WER increase compared to full encoder fine-tuning. Residual adapters, while using less than 0.2\% of the parameters, perform significantly better than updating only the first encoder layer (9\% of parameters on RNN-T) and slightly better than updating the encoder layers 1-3 (32\% of parameters on RNN-T).

For accented speech, Table~\ref{tab:summary_results} shows that fine-tuning leads to more moderate improvements of 29\% for RNN-T and 35\% for T-T (averaged across all accents). Similar to the personalization task, residual adapters performs slightly worse than fine-tuning (24\% improvement for RNN-T and 31\% for T-T), but require the update of only a fraction of parameters. The alternative of updating only the first encoder layer shows a much poorer performance.

Figure~\ref{fig:adapter_loss_combined} shows the \textit{adapter performance drop}, the relative WER reduction
when switching from fine-tuning to residual adapters calculated per speaker/accent (lower is better):
\begin{equation}
\label{eq:delta}
    \delta = \frac{\textrm{WER}(\textrm{adapters}) - \textrm{WER}(\textrm{fine-tuned})}{\textrm{WER}(\textrm{fine-tuned})}    
\end{equation}
T-T exhibits a lower average adapter performance drop $\delta$  compared to RNN-T on both tasks. This is likely due to the higher overall capacity of residual adapters when applied to T-T due to the higher number of encoder layers (15 encoder layers for T-T, 8 for RNN-T).

\subsection{Hyper-Parameter Tuning}
\label{ss:hparam_tuning}

The results reported in Table~\ref{tab:summary_results} are for bottleneck dimension and learning rates found to work well in hyper-parameter tuning experiments where we ran a grid search over a combination of the two. For the learning rate, we evaluated ($1e-5$, $1e-4$, $1e-3$, $1e-2$), and for the bottleneck dimension, we evaluated $4,16,32,128$.
We use a random subset of 20 speakers for the atypical speech task for parameter tuning 
to make search feasible; for accents, search was run across all 10 accents.

For the atypical speech personalization task, a learning rate of $1e-5$ worked best for fine-tuning, and $1e-3$ for residual adapters (both on RNN-T and T-T).\footnote{For individual speakers, higher or lower learning rates might individually lead to better performance. However, in a practical personalization scenario  with hundreds of speakers, such tuning is impossible, so we chose a one-size-fits-all approach.}
For accents, we found the best learning rate for fine-tuned models using RNN-T to be $1e-5$, and $1e-4$ for T-T. For residual adapters, the best learning rate was $1e-4$ (RNN-T) and $1e-3$ (T-T). 
We found that accents with high amounts of training data tended to be more tolerant to higher learning rates compared to accents with limited training data.
 
Overall, these experiments show that adapters require on average a learning rate one order of magnitude higher than fine-tuning the encoder for both T-T and RNN-T. This may be attributable to the much smaller capacity of adapters and/or to their random initialization.

For the bottleneck dimension, we found that $b_d=4$ often leads to a  higher adapter performance drop on the atypical speech task, although for some speakers -- especially those with mild impairment and generally relatively low (<= 25) WER on the unadapted models -- even this bottleneck dimension worked very well. A bottleneck dimension of $b_d=128$ rarely led to increased performance over $b_d=16$ and $b_d=32$. Between the latter two, we could not make out a clear pattern; they often performed equally well. 
For accented speech, when adapting for individual accents, we found that both $b_d=4$ and $b_d=16$ achieve similar performance.
A bottleneck dimension $b_d=128$ on average leads to worse performance than $b_d=16$.
Given these results, we chose a bottleneck dimension of $b_d=16$ for all reported experiments, unless otherwise noted.

\subsection{Atypical Speech Personalization}

In this section we further analyze performance of the residual adapters for the atypical speech personalization task, zooming in on aspects like speaker impairment severity and phrase types.

\label{sec:exp_atypical_severity}

\begin{table*}[h]
  \centering
  \begin{tabular}{|l|r|r|r|r|r|r|} \hline
  \multirow{2}{*}{Adaptation style}  & \multicolumn{3}{|c|}{RNN-T} & \multicolumn{3}{|c|}{T-T} \\
                    & mild & moderate & severe & mild & moderate & severe  \\ 
  \hline
  Unadapted           & 20.2  & 44.2  & 78.1 & 16.5 & 42.9 & 76.9 \\
  Fine-tune, full enc &  4.1 (76\%) &  6.5 (84\%) & 14.2 (79\%) &  4.5 (74\%) &  6.9 (83\%) & 13.3 (78\%) \\
  Residual Adapters         &  4.8 (70\%) &  7.5 (80\%) & 16.3 (77\%) &  4.8 (71\%) &  8.2 (81\%) & 15.3 (77\%) \\
  \hline
  \end{tabular}  
  \caption{Results for atypical speech personalization task, broken down by severity (reported: median WER scores with relative WER improvement over the unadapted base model in brackets).}
  \label{tab:impaired_speech_by_severity}
\end{table*}

\begin{figure}[b]
    \centering
    \includegraphics[width=0.47\textwidth]{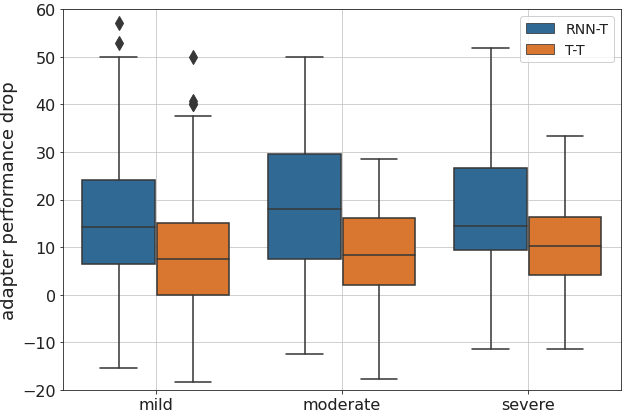}
    \caption{Distribution of adapter performance drop by severity of speech impairment.}
    \label{fig:adapter_loss_by_severity}
\end{figure}

Figure~\ref{fig:adapter_loss_by_severity} shows the adapter performance drop $\delta$ per severity. For T-T, residual adapters seem to work similarly well across all 3 types of severities. 
On RNN-T models, we observe that $\delta$ is higher for speakers with moderate severity (median decrease of 18\% for moderate vs 14\% for mild and severe). In particular, we found that $\delta$ is somewhat correlated with the relative WER improvement $\gamma$ of the fine-tuned model: $\delta$ is higher for cases where the adaptation by fine-tuning helps the most (Spearman correlation coefficient of 0.344 ($p < 0.001$) for RNN-T, more moderate correlation of 0.189 (p=0.06) for T-T). 
Overall, adaptation by fine-tuning and residual adapters show similar behavior across severity levels, which suggests that residual adapters do not have disadvantages for specific severity levels.

\begin{table*}[t]
  \centering
  \begin{tabular}{|l|r|r|r|r|r|r|} \hline
  \multirow{2}{*}{Adaptation style} &  \multicolumn{3}{|c|}{home automation} & \multicolumn{3}{|c|}{conversational} \\ 
                   &  mild & moderate & severe &  mild & moderate & severe  \\
  \hline  
  Unadapted            & 17.4 & 41.6 & 61.1 & 23.7 & 39.1 & 82.4 \\
  Fine-tune, full enc  & 3.7 (76\%) & 3.6 (88\%) & 7.8 (85\%) & 6.2 (67\%) & 4.9 (83\%) & 12.4 (71\%) \\
  Residual Adapters          & 4.7 (76\%) & 5.0 (87\%) & 8.5 (84\%) & 6.0 (65\%) & 5.8 (79\%) & 15.0 (72\%) \\
  \hline
  \end{tabular}
  \caption{Results for atypical speech personalization task, broken down by domain (reported: median WER scores with relative WER improvement over the unadapted base model in brackets). Results are median scores over all speakers per severity group. We show results for T-T only due to space constraints; RNN-T results are analogous.}
  \label{tab:impaired_speech_by_domain}
\end{table*}

The Euphonia corpus also comes with domain information for each utterance. 
This enables us to analyze the performance of fine-tuning and residual adapters on two different domains -- home automation queries\footnote{Examples: "turn on lights" or "play ABBA on Spotify".} (short phrases of  3.2 words on average) and conversational phrases (longer with  7.4 words on average, open domain) -- to understand whether residual adapters have
trouble with different phrase types.
Table~\ref{tab:impaired_speech_by_domain} shows T-T WERs for these two domains for a subset of 43 speakers who had a sufficient number of recordings of both home automation and conversational phrases. The conversational domain, being longer and with a more open vocabulary, generally is more challenging for ASR and accordingly across all severity levels we observe higher WERs. Moreover, this domain seems to be harder to adapt to, resulting in lower WER improvements through both types of adaptation, most notably on the severe group where adaptation gain drops from $\sim85\%$ to  $\sim71\%$ (fine-tuning approach).  Despite these difference, both fine-tuning and residual adapters show similar behavior, and we conclude that even for more challenging domains with longer utterances, residual adapters work well.

\subsection{Accent Adaptation}

\begin{table*}[t]
  \centering
  \begin{tabular}{|l|l|l|c|c|c|c|c|c|c|c|c|c|c|} \cline{2-13} 
  \multicolumn{1}{c}{} & \multicolumn{2}{|l|}{Accents} & \texttt{af} & \texttt{au} & \texttt{ca} & \texttt{en} & \texttt{in} & \texttt{ir} & \texttt{nz} & \texttt{ph} & \texttt{sc} & \texttt{si} \\ \cline{2-13}
  \multicolumn{1}{c|}{} & \multirow{2}{*}{\# utt} &  train & 1.6k & 17.7k & 13.2k & 31k & 20.1k & 2.9k & 2.3k & 1k & 4.4k & 1.2k\\ 
  \multicolumn{1}{c|}{} & & dev/test & 300 & 2.2k & 1.7k & 3k & 2.5k & 360 & 300 & 300 & 550 & 300\\ \hline
  \multirow{5}{*}{\rotatebox{90}{RNN-T}} & \multicolumn{2}{|l|}{Unadapted} & 16.1 & 17.3 & 11.5 & 13.7 & 20.0 & 11.6 & 13.4 & 18.3 & 56.3 & 20.5 \\
  \cline{2-13}
  & \multicolumn{2}{|l|}{Per-Accent, Fine-tune} & 11.4 & 11.1 & 9.7 & 10.6 & 13.8 & 9.8 & 9.0 & 12.8 & 28.0 & 14.8 \\
  & \multicolumn{2}{|l|}{Per-Accent, Res Adapt} &  11.8 & 12.3 & 10.3 & 11.0 & 15.3 & 10.1 & 10.1 & 13.9 & 30.9 & 15.6 \\ \cline{2-13}
  & \multicolumn{2}{|l|}{Multi-Acc, Fine-tune} & 11.7 & 12.6 & 10.4 & 11.6 & 15.5 & 9.8 & 9.4 & 12.3 & 30.6 & 15 \\ 
  & \multicolumn{2}{|l|}{Multi-Acc, Res Adapt} & 12.2 & 13.3 & 10.5 & 12.1 & 16.9 & 10.4 & 10.9 & 14.1 & 33.6 & 16.3 \\ 
  
  \hline \hline
  \multirow{5}{*}{\rotatebox{90}{T-T}} & \multicolumn{2}{|l|}{Unadapted} & 16 & 18.9 & 13.2 & 15.3 & 20.8 & 13.7 & 13 & 19.9 & 61.2 & 24.2 \\
  \cline{2-13}
  & \multicolumn{2}{|l|}{Per-Accent, Fine-tune} & 11.8 & 11.5 & 9.6 & 10.8 & 13.8 & 9.9 & 9.0 & 12.3 & 29.2 & 13.9 \\
  & \multicolumn{2}{|l|}{Per-Accent, Res Adapt} &  12.5 & 12.0 & 9.8 & 11.1 & 14.7 & 9.9 & 10.3 & 13.5 & 31.9 & 15.3 \\ \cline{2-13}
  & \multicolumn{2}{|l|}{Multi-Acc, Fine-tune} & 11.2 & 12.8 & 9.7 & 11.3 & 15.2 & 9.5 & 9.5 & 11.9 & 33.5 & 14.3 \\ 
  & \multicolumn{2}{|l|}{Multi-Acc, Res Adapt } & 12.7 & 14.1 & 10.3 & 11.7 & 16.2 & 10.7 & 10.1 & 13.1 & 36.1 & 14.8 \\ 
  
  \hline
  \end{tabular}
  \caption{WER on the accent adaptation task, showing fine-tuning (full encoder) and residual adapters for the per-accent and multi-accent adaptation scenarios.}
  \label{tab:accents_both}
\end{table*}

Table \ref{tab:accents_both} presents the per-accent WER results.
Depending on the accent and the amount of training data, we observe substantial variance with respect to WER of the unadapted and adapted models. 
Across all accents, the Scottish accent (\texttt{sc}) performs worst with extremely high WER, both for the unadapted and adapted models.\footnote{Note that beyond accent variation, results shown in Table~\ref{tab:accents_both} are affected by a domain mismatch between the training data used for the base model and the Common Voice corpus.}
However, even for accents with fairly small amounts of training data, such as \texttt{af}, adaptation clearly improves the performance over the unadapted model. 

While fine-tuning the full encoder has slightly better performance than residual adapters, the adapter performance drop $\delta$ is relatively small (see Figure~\ref{fig:adapter_loss_combined}). This is consistent with our findings on the atypical speech personalization task (Section \ref{sec:exp_atypical_severity}), where lower relative WER improvement is associated with much lower adapter performance drop.
Analogously to the atypical speech personalization task, the adapter performance drop is smaller on T-T compared to RNN-T.

In order to provide a better context to the related work on accent recognition and to test the capability of the residual adapters in large group adaptation scenarios, we also ran experiments for  multi-accent adaptation where a single model is fine-tuned (full encoder update) or adapted with residual adapters to all accents at once. We increased the bottleneck size to $b_d=128$ so that the residual adapters have a sufficiently large capacity to handle this more complex task.
A balanced training set of 11 accents (10 accents as described in Section \ref{sec:accent_data} plus the US accent; 10k utterances per accent, up- or down- sampled depending on the amount of the training data per accent) was used for adaptation. Similarly, a balanced dev set was used to identify the best performing checkpoint. For WER calculation, we used the original test set per accent for comparability. 

Results in Table~\ref{tab:accents_both} show that even in a large group adaptation scenario, residual adapters perform well and their performance is comparable to the per-accent adaptation scenario (adapter performance drop $\delta \approx 8\%$ for both RNN-T and T-T).

\begin{table}[t]
  \centering
  \begin{tabular}{|l|c|c|c|c|}
  \hline
   & \multicolumn{2}{|c|}{best checkpoint} & \multicolumn{2}{|c|}{global steps/sec} \\
  Arch & Fine- & Residual & Fine- & Residual \\
       & tune & Adapters  & tune  & Adapters \\ 
  \hline        
  RNN-T & 20380 & 5610 & ~1.3 & ~1.8 \\ 
  T-T   & 5150 & 5300 &  ~3.6 & ~5.2\\
  \hline
  \end{tabular}
  \caption{Median number of steps for best checkpoint and global steps/second for the atypical speech personalization task showing that residual adapters lead to about 40\% reduction in training time.}
  \label{tab:training_time_analysis}
\end{table}
\subsection{Training and inference time}

On the atypical speech personalization task, when updating residual adapter parameters only -- as opposed to encoder fine-tuning -- we observed a speedup in training time as measured in global steps per second.
Table~\ref{tab:training_time_analysis} compares the median number of training steps needed across the 100 speakers (i.e. best checkpoint selected) as well as the global steps/second for fine-tuning the full encoder vs residual adapter training. While residual adapters led to about the same (T-T) or fewer (RNN-T) number of steps for convergence, the global steps/second score increased by about 40\%.\footnote{We used the same accelerator setup (2x2 Tensor Processing Units slices) for fine-tuning and residual adapter training.}
This gain is especially relevant for a personalization scenario where large numbers of user-specific models need to be trained. 

During inference, on the other hand, we didn't observe a measurable increase in latency when adding residual adapters. To test this, we decoded test sets of around 300 utterances several times on personalized models trained with and without residual adapters. 
\section{Conclusions and Future Work}
\label{sec:conclusions}

In this work, we have shown that adaptation of ASR models using residual adapter layers leads to substantial WER improvements over unadapted models across two tasks: atypical speech (up to 77\% relative WER reduction) and accented speech (up to 31\% relative reduction) and in two  architectures (RNN-T and T-T). In comparison, fine-tuning the entire encoder for each speaker or accent yields only small improvements compared to residual adapter training. 

While similar in adaptation performance, residual adapters are much more parameter efficient than model fine-tuning. In our scenario, using residual adapters on each encoder layer, less than 0.5\% of the overall model parameters need to be trained and maintained per speaker or accent. On the other hand, fine-tuning the entire encoder affects over 80\% of the model parameters.
In addition to substantially improved parameter efficiency, we also observed a dramatic training time speed up of about 40\% due to the reduced number of parameter updates.

Overall, these findings demonstrate a feasible and scalable solution for personalized, speaker-dependent models as well as domain-specific or dialect/accent-focused models.

In future work, we plan to study to which encoder layers we need to add adapters for best performance and to potentially make residual adapters even more parameter efficient. Similarly, we plan to apply residual adapters with different bottleneck dimensions depending on the position in the encoder layer stack (bottom and middle layers likely require larger, top layers smaller capacity).
Finally, we also plan to directly compare the effectiveness of residual adapters to approaches using statically fed speaker-dependent vectors for speaker adaptation, especially in the context of accent adaptation.

\bibliography{anthology,custom}
\bibliographystyle{acl_natbib}

\end{document}